\documentclass[letterpaper, 10 pt, conference]{ieeeconf}  

\IEEEoverridecommandlockouts                              

\usepackage{graphics} 
\usepackage{epsfig} 

\title{\LARGE \bf
Integrating and Evaluating Visuo-tactile Sensing with Haptic Feedback for Teleoperated Robot Manipulation
}

\author{
Noah Becker$^{1}$,
Kyrylo Sovailo$^{1}$,
Chunyao Zhu$^{1}$,
Erik Gattung$^{1}$,\\
Kay Hansel$^{1}$,
Tim Schneider$^{1}$,
Yaonan Zhu$^{5}$,
Yasuhisa Hasegawa$^{5}$,
and Jan Peters$^{1-4}$
\thanks{
This research received funding from CHIRON by DFG (PE 2315/8-1), the European Union’s Horizon Europe programme under Grant Agreement No. 101135959 (project ARISE), BMWSB ZukunftBau (no. 10.08.18.7-21.34), and Aristotle by BMBF;
$^{1}$Computer Science Department, Technische Universität Darmstadt, Germany,
$^2$German Research Center for AI (DFKI), 
$^3$Hessian Centre for Artificial Intelligence (Hessian.AI), 
$^4$Centre for Cognitive Science (CogSci). {\texttt{\{noah.becker, erik.gattung\}  @stud.tu-darmstadt.de}, \texttt{\{kay.hansel, tim.schneider1, jan.peters\} @tu-darmstadt.de}}, \newline
$^5$Department of Micro-Nano Mechanical Science and Engineering,
Nagoya University, Japan. {\texttt{zhu@robo.mein.nagoya-u.ac.jp}, \texttt{hasegawa@mein.nagoya-u.ac.jp}};
}%
}

\begin{document}

\maketitle
\thispagestyle{empty}
\pagestyle{empty}
\begin{abstract}
Telerobotics enables humans to overcome spatial constraints and physically interact with the environment in remote locations.
However, the sensory feedback provided by the system to the user is often purely visual, limiting the user's dexterity in manipulation tasks.
This work addresses this issue by equipping the robot's end-effector with high-resolution visuotactile GelSight sensors.
Using low-cost MANUS-Gloves, we provide the user with haptic feedback about forces acting at the points of contact in the form of vibration signals.
We employ two different methods for estimating these forces; one based on estimating the movement of markers on the sensor surface and one deep-learning approach. 
Additionally, we integrate our system into a virtual-reality teleoperation pipeline in which a human user controls both arms of a Tiago robot while receiving visual and haptic feedback.
Lastly, we present a novel setup to evaluate normal force, shear force, and slip.
We believe that integrating haptic feedback is a crucial step towards dexterous manipulation in teleoperated robotic systems.
\end{abstract}

\section{INTRODUCTION}
Teleoperation, the remote control of robots, enables humans to overcome spatial constraints and physically interact with the environment in remote locations~\cite{sheridan1992telerobotics,sheridan1995teleoperation}. 
Recently, contributions such as the Nimbro system~\cite{schwarz2021nimbro,lenz2023nimbro}, the ALOHA system~\cite{zhao2023learning}, or the mobile ALOHA system~\cite{fu2024mobile} attracted a lot of attention in the field.
A common challenge lies in providing feedback that is not easily conveyed through images, like temperature, surface structure, or grip force. 
Haptic feedback has been considered as a solution to this problem, providing additional information through touch instead of sight~\cite{muradore2016review, niemeyer2016telerobotics}.
This information is particularly relevant in our context, as we aim for an intuitive teleoperation system designed for people with limited technical knowledge. 

\begin{figure}[t!]
    \centering
    \includegraphics[width=1.0\linewidth]{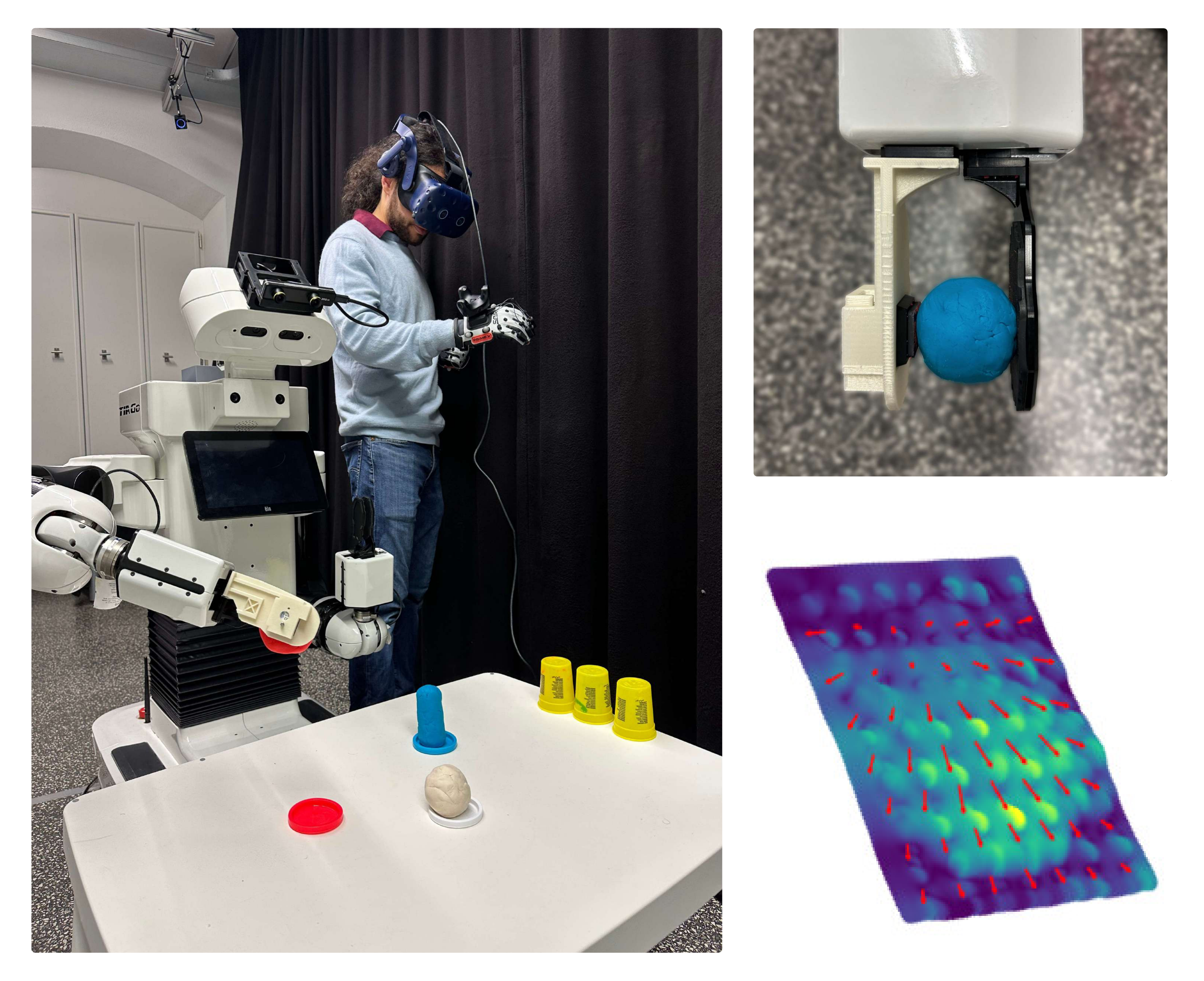}
    \vspace{-1.0em}
    \caption{The Teleoperation setup. \textbf{Left - }Tiago robot equipped with tactile GelSight sensor. \textbf{Top right - }Grasping modeling compound, with GelSight in white finger. \textbf{Bottom right - } Visualisation of motion flow used for force estimation.}
    \label{fig:cover}
    \vspace{-1.5em}
\end{figure}

\begin{figure*}[!t]
    \centering
    \includegraphics[width=1.0\linewidth]{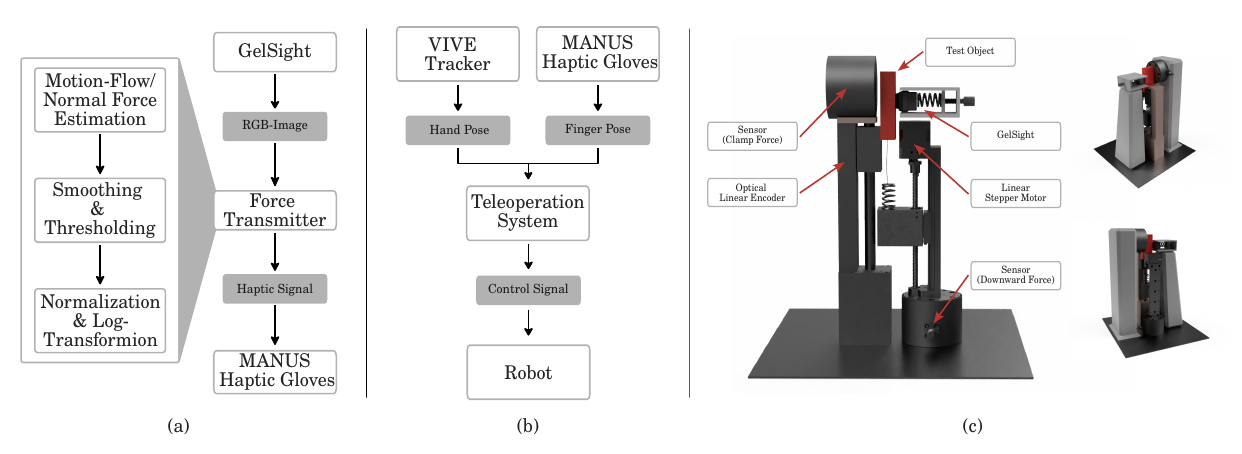}
    \vspace{-2.0em}
    \caption{Overview of different components of the proposed system. 
    \textbf{(a) - } The haptic feedback pipeline. 
    \textbf{(b) - } The teleoperation pipeline. 
    \textbf{(c) - } The test setup for determining the slip and shear forces. 
    \textbf{(c) left - } Rearranged for better visibility of the individual components.\textbf{(c) right - } Our complete test setup including casing.
    } 
    \label{fig:overview-pipeline}
    \vspace{-1.5em}
\end{figure*}

In this work, we integrated vibrotactile feedback into a virtual reality (VR) teleoperation pipeline to improve performance in object manipulation tasks. 
This addition extends the visual feedback received from the VR system by providing tactile sensations to the user.
The user operates a dual-arm manipulation robot, benefiting from both visual and tactile cues to improve control and precision.
Vibration feedback, also called vibrotactile feedback, is one of the most commonly used types of tactile feedback~\cite{martinez2014identifying, weber2016daglove, taichi_2016_teleop}; used due to its low profile, cost, and power consumption. 
We attached visuotactile sensors to the robot's grippers to detect force exertion, convert it to vibrotactile feedback, and transmit this in real-time to the user for intuitive control.
While gripping force could also be measured with traditional force sensors, visuotactile sensors can provide much richer information about the points of contact, such as object texture~\cite{böhm2024matters,Yuan2018May}, object shape~\cite{suresh2022shapemap}, or the presence of slip~\cite{funk2023evetac}. 
Lastly, we introduce a novel test setup for precise quantitative analysis and potential data collection of normal forces, shear forces, and slip.

\section{SYSTEM AND METHOD}
We integrate haptic feedback into a VR teleoperation pipeline, as shown in Figure \ref{fig:overview-pipeline}. 
The user wears an HTC Vive VR headset~\cite{vive} providing a first-person view through a ZED Mini stereo camera~\cite{zed_mini} mounted on the robot's head.
This setup enables intuitive control of a dual-arm manipulation robot such as the Tiago~\cite{tiago}, akin to systems described in works \cite{zhu2023visual, zhu2023shared, becker_2024_visuoteleop}. 
For haptic feedback, we use two Prime X haptic gloves from MANUS~\cite{manus}, equipped with flex sensors and vibration motors for each finger.
The flex sensors detect the user's finger pose, a gesture that the robot's grippers mimic. 
Additionally, a tracker on the glove tracks the user's hand pose and enables the robot to replicate arm motions. Communication between system components is done using the Robot Operating System (ROS)~\cite{ros} and shared memory to minimize latency.

The haptic feedback pipeline uses a GelSight Mini \cite{gelsight,gelsight_shear_force} with a dot-matrix-gel mounted to the robot's gripper to sense contact with an object.
The GelSight streams the video of its gel to a transmitter node, which converts the images to a haptic signal.
This haptic signal gets forwarded to the gloves as vibration feedback.
To generate the haptic signal, we employed the Lucas-Kanade optical motion flow method \cite{lucas-kanade} and a neural network \cite{shear-force,funk_2023_shear_force} trained to estimate the force acting on the gel.
As motion flow is an adequate approximation of force~\cite{gelsight,gelsight_shear_force}, we 
primarily used the former deterministic method. We apply an additional threshold to the total estimated force calculated by either algorithm to prevent the glove from vibrating due to noise or arm movements. 
In addition, the values are normalized, and a log scale is applied so that the haptic feedback range for smaller forces is larger, assisting with fine control.
The resulting force estimates are sent to the MANUS glove, which controls the haptic motors to vibrate proportional to the total force.

\section{EVALUATION}

We conducted a small preliminary user study with seven participants to evaluate our system's strengths and weaknesses. 
In the study, participants were tasked with using our teleoperation system to pick up a plasticine ball as gently as possible without dropping it (see Figure \ref{fig:cover}). Balls of different sizes were employed, and the haptic feedback was either enabled or disabled.
The preliminary results show that, on average, ball deformation was reduced by 48\% when haptic feedback was enabled.
For the qualitative analysis, we use the NASA Task Load Index (TLX) \cite{NASA-TLX}, which indicates that users perceive haptic feedback as beneficial for enhancing performance.
However, we also noticed that performing fine control with the increased feedback requires additional effort, mostly due to the system's high sensitivity requiring careful and fine movement. 
Further user studies, both on quantitative and qualitative measurements, are still ongoing.

Furthermore, we introduce a novel test setup for precise quantitative analysis of shear force and slip (see Figure \ref{fig:overview-pipeline}).
In the setup, a test object is clamped between a GelSight sensor and a force sensor, which measures the normal force acting on the GelSight through the test object.
A linear stepper motor -- mounted on a second force sensor -- is attached to the test object through a spring and a wire. 
This second sensor measures the shear force acting on the GelSight through the test object.

With the motor, we can induce a controlled shear force between the test object and the GelSight by tensioning the spring.
We can detect the occurrence of slip by applying and subsequently releasing tension to the test object and measuring whether its position changed at the end of the procedure.
If no slip occurred, we increase the tension and repeat the procedure. 
Because of the short time frame in which the system is tensioned, we can accurately determine the time and force at which the slip occurred.
 
\section{Conclusion}
In this work, we developed a teleoperation system where users get haptic force feedback through vibration. 
We conducted a preliminary user study and introduced a novel test setup, which will be used to evaluate future visual-tactile-based methods to integrate into our system.
In future work, we additionally plan to utilize the rich data of visuotactile sensors to provide additional haptic feedback to the user, such as shear forces, slip, and texture, further reducing the reliance on visual cues.
We currently use the test setup to benchmark different slip-detection methods \cite{Yuan,Dong,Li2015TouchingIB,luo2023beyond}. 
In the future, the data collected during these experiments might be used for machine learning applications.
Another exciting avenue for further research is to utilize these sensors in shared control, e.g., by automatically adjusting gripping strength based on tactile feedback.

\bibliographystyle{IEEEtran}
\bibliography{main}

\end{document}